\documentclass[conference]{IEEEtran}
\IEEEoverridecommandlockouts

\usepackage{cite}
\usepackage{amsmath,amssymb,amsfonts}
\usepackage{algorithmic}
\usepackage{graphicx}
\usepackage{textcomp}
\usepackage{xcolor}
\usepackage{comment}
\usepackage{multirow}
\usepackage{booktabs}
\usepackage{siunitx} 
\usepackage{array}
\usepackage{makecell}
\usepackage{float}
\usepackage{balance}

\def\BibTeX{{\rm B\kern-.05em{\sc i\kern-.025em b}\kern-.08em
    T\kern-.1667em\lower.7ex\hbox{E}\kern-.125emX}}

\makeatletter
\newcommand{\linebreakand}{%
 \end{@IEEEauthorhalign}
 \hfill\mbox{}\par
 \mbox{}\hfill\begin{@IEEEauthorhalign}
}
\makeatother

\begin{document}

\title{ParticleSAM: Small Particle Segmentation for Material Quality Monitoring in Recycling Processes
\thanks{This work was partially funded by the German Ministry of Education and Research (BMBF)
under Grant Agreements 02WDG1693D (KIMBA) and
033R390(ReVise\_UP).}\\


}

\author{
    Yu Zhou\IEEEauthorrefmark{1}, Pelle Thielmann\IEEEauthorrefmark{1}, Ayush Chamoli\IEEEauthorrefmark{1}, Bruno Mirbach\IEEEauthorrefmark{1}, Didier Stricker\IEEEauthorrefmark{1}\IEEEauthorrefmark{2}, Jason Rambach\IEEEauthorrefmark{1}\\
    \IEEEauthorrefmark{1}German Research Center for Artificial Intelligence (DFKI), 
    \IEEEauthorrefmark{2}RPTU Kaiserslautern\\
    \{yu.zhou, pelle.thielmann, bruno.mirbach, didier.stricker, jason.rambach\}@dfki.de, chamoli@rptu.de
}
\maketitle

\begin{abstract}
The construction industry represents a major sector in terms of resource consumption. Recycled construction material has high reuse potential, but quality monitoring of the aggregates is typically still performed with manual methods. Vision-based machine learning methods could offer a faster and more efficient solution to this problem, but existing segmentation methods are by design not directly applicable to images with hundreds of small particles.  
In this paper, we propose ParticleSAM, an adaptation of the segmentation foundation model to images with small and dense objects such as the ones often encountered in construction material particles. 
Moreover, we create a new dense multi-particle dataset simulated from isolated particle images with the assistance of an automated data generation and labeling pipeline. This dataset serves as a benchmark for visual material quality control automation while our segmentation approach has the potential to be valuable in application areas beyond construction where small-particle segmentation is needed. Our experimental results validate the advantages of our method by comparing to the original SAM method both in quantitative and qualitative experiments. 
\end{abstract}

\section{Introduction}
The construction industry is known for its high demand for raw materials, particularly aggregates such as sand, gravel, and crushed stone. According to the European Aggregates Association (UEPG)\cite{aggregatesEurope}, the European aggregates demand is 3 billion tonnes annually. This has led to resource shortages and environmental impacts that urgently need to be addressed and resolved.

Increasing the use of recycled (RC) aggregates as a substitute for primary aggregates, particularly in the construction industry, is a cornerstone tactic for minimizing the environmental impact and achieving sustainability goals. 
Traditional characterization methods of recycled Construction and Demolition Waste (CDW), which often rely on manual sorting and expert visual inspection, yield delayed results and are prone to human error.
The integration of image-based monitoring technologies into the production and quality inspection process of CDW has the potential to replace the labor-intensive and time-consuming manual analysis methods. This can be applied in tasks such as the estimation of the Particle Size Distribution (PSD) to reduce the cycle time of qualified RC aggregates and enhance their overall quality.  

Currently, the development and application of deep learning models in the field of CDW analysis face challenges due to the lack of publicly available open-source RC aggregates dataset. Such datasets are crucial for the training, validation, and evaluation of deep learning models. Moreover, while state-of-the-art deep learning models have demonstrated remarkable performance in standard instance segmentation tasks, their effectiveness diminishes when applied to industrial images, which are typically high resolution and often contain multi-layers and densely occluded small particles.

To address these challenges, we build upon the Segment Anything Model (SAM)\cite{kirillov2023segment} and propose necessary modifications to segment large numbers of small-sized particles from industrial material flow images with improved accuracy and efficiency. 
To encourage research in the topic, we build a benchmark high-resolution multi-layer crushed CDW particle dataset simulated from isolated particle images. Our proposed data generation engine automatically segments isolated particle instances from industrial single-layer images, refines them, and simulates augmented multi-particle images. 

 To summarize our main contributions:
 \begin{itemize}
     \item We generate a benchmark image dataset of multi-layer CDW particles at varying overlap levels and particle size distributions.
     \item We propose ParticleSAM, an adaptation of the Segment Anything Model (SAM) for segmenting images of densely placed small particles, tested for CDW, but applicable to any image dataset of similar characteristics.
 \end{itemize}

\section{Related Work}

\begin{figure*}[t]
    \centering
    \includegraphics[width=0.8\textwidth]{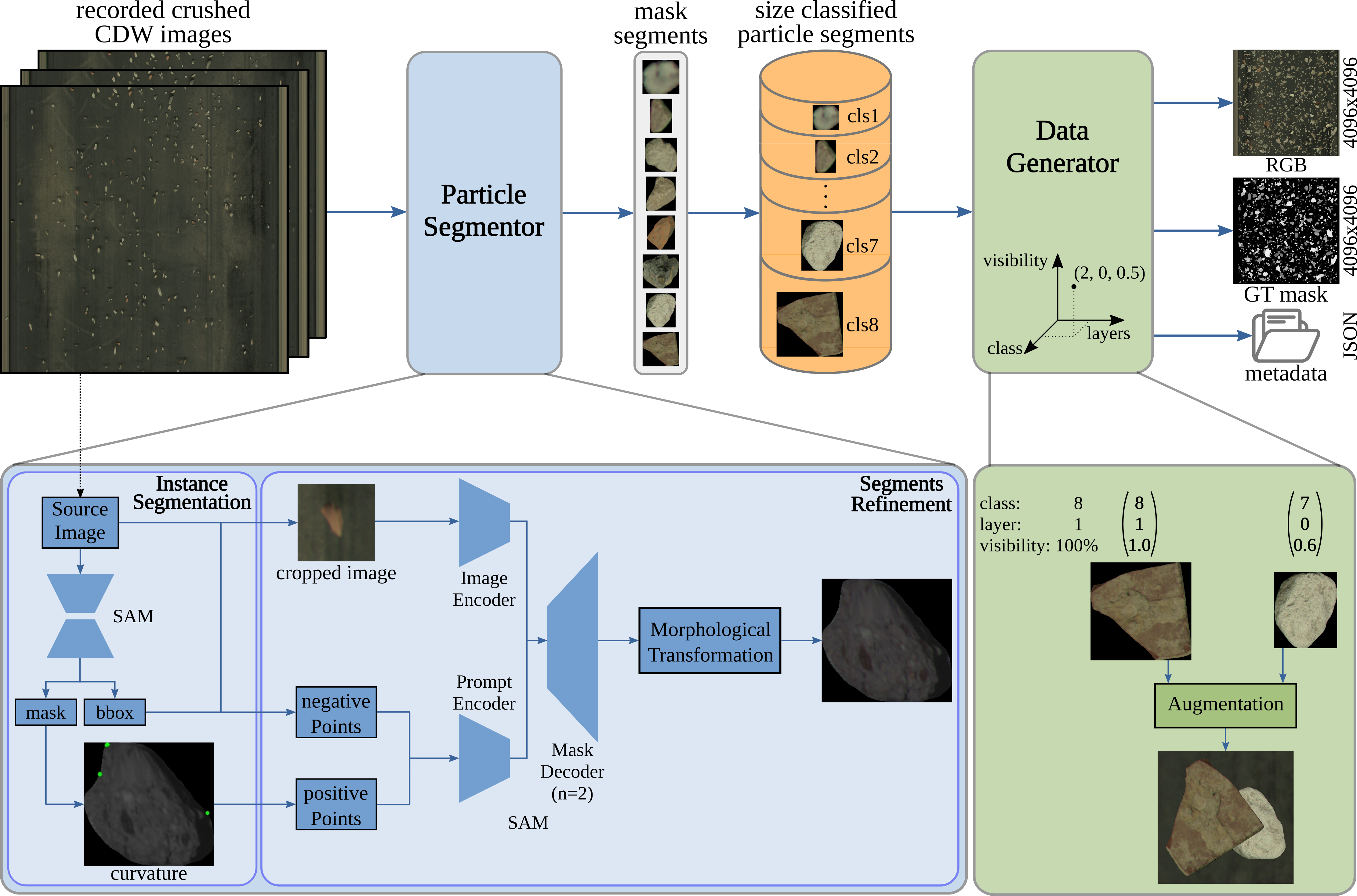}  
    \caption{Data Simulation Pipeline: The Particle Segmentor extracts isolated particles from conveyor belt recordings using SAM twice. Segmented particles are classified by size and stored for the Data Generator. The Data Generator augments particles based on class, layer, and visibility, producing high-resolution RGB images, ground truth masks, and metadata.
    \textit{The recorded crushed CDW image provided by Lieve Göbbels from the department of Anthropogenic Material Cycles (ANTS), RWTH Aachen University.}}
    \label{Figure 1}
\end{figure*}

\subsection{Sensor-based Characterization of Construction Waste}\label{AA}

Recently, sensor-based approaches have gained traction in CDW recycling, utilizing imaging, spectroscopy, and machine learning to improve the analysis of material characterization.

Near-Infrared (NIR) camera-based approach can achieve a real-time monitoring of CDW materials \cite{hernandez2021near}, excelling at material differentiation and performing well in low-light environments. However, NIR provides very low resolution, making it incapable of small particle size prediction.  
3D laser triangulation (3DLT) sensor \cite{dlPSD} and fusion of RGB and depth cameras \cite{rgbd} are integrated to acquire additional depth information except 2D data. They are combined with CNN-based classification and segmentation models, respectively. By fusing the depth information, the detection of particle size can be optimized during monitoring; however, even in a stable laboratory environment, these systems require highly precise and complex installation and calibration processes. 

Methods focusing solely on RGB inputs have demonstrated that competitive performance can also be achieved by combining deep learning technologies. As in \cite{CDWdatasets}, they contributed an open access CDW dataset as the basis for training classification models. In \cite{doadaptSeg} and \cite{wastesam}, segmentation models are modified 
to provide detailed boundary and shape information, enabling more precise analysis than object detection. Nevertheless, both studies focus on initial solid construction waste, with limited contribution to the recyclability and reusability of aggregates derived from crushed CDW. In \cite{kronenwett2024sensor}, the authors first introduce a synthetic datasets composed of real brick and sand-lime samples, and then re-train CNN detection models on this synthetic data to enhance their accuracy in domain-specific object detection tasks. Even so, the very low resolution (300 pixels) of their data renders it completely unsuitable for small particle segmentation.

\subsection{Segmentation Foundation Model}

Foundation models have advanced the field of instance segmentation by training on massive large-scale datasets and enabling them adaptable to various downstream tasks with minimal fine-tuning. SAM \cite{kirillov2023segment} and Mask2Former \cite{mask2former} both employ the state-of-the-art vision transformer-based encoder-decoder architecture with attention mechanisms to achieve flexible and robust instance segmentation. Moreover, SAM is pre-trained on a vastly larger dataset compared to Mask2Former.
Recent works have explored integrating SAM with downstream segmentation tasks. For example, researchers have fine-tuned SAM with weak annotations \cite{das2024prompting}, using its output to generate high-quality pseudo-labels for training lightweight instance segmentation models. Med-SA\cite{wu2023medical} is a specialized adaptation of SAM, fine-tuned on medical imaging datasets, it adapts 2D SAM to 3D medical images to achieve higher precision in medical image segmentation. 

SAM is the first general foundation model for instance segmentation, capable of zero-shot transfer to a wide range of downstream tasks. Our research focuses on adapting the SAM model to enhance its segmentation capability for CDW images, addressing a critical challenge in sensor-based analysis of recycled aggregates.


\section{Methodology} 
\subsection{Data Simulation Pipeline}

Up to now, research conducted on CDW datasets \cite{CDWreview} has mainly focused on raw CDW, which serves as the input material for processing plants during recycling. In contrast, our study shifts the focus to processed CDW aggregates, which undergo particle size sorting before being distributed to construction industries as RC materials. As mentioned above, there is currently no such open source dataset. Kronenwett et al.~\cite{kronenwett2024sensor} represents a first step in this direction, but their dataset is limited in particle size and type diversity and contains low-resolution synthetic images. 

Our idea has been to generate augmented multi-particle images with varying overlap and size distribution. 
Fig.~\ref{Figure 1} shows the full data simulation pipeline. We first extract isolated particles from single-layer image recordings with a segmentor, and then use a generator to automatically output augmented images after a size classification.

\textbf{Particle Segmentor}: Inspired by SAM’s data engine, we design a Particle Segmentor (see Fig.~\ref{Figure 1}), which reconfigures the two encoders and decoders of SAM to achieve automatic instance segmentation and segments optimization.
Initially, we segment particles from images of crushed CDW, which were captured while the aggregates were being transported on a conveyor belt.
In industry environments where the lighting conditions are often suboptimal, segmentation faces significant challenges due to shadows and motion blur. To address these issues, we add a refinement module: corner points from bounding box and curvature points from the mask output are extracted as negative and positive points, respectively, serving as input of the prompt encoder. Subsequently, morphological transformations are applied to optimize segmentation masks by removing noise, filling gaps, smoothing edges, and connecting broken regions.

\textbf{Particle size Classification}: According to the manual sieve analysis defined by DIN 66165-1 suggested by \cite{DIN}, we classify the obtained particles into 8 classes from 4 \text{mm} to 63 \text{mm} based on their farthest pair distance (see Table~\ref{tab:class_layer}).

\textbf{Data Generator}: To facilitate the efficient generation of large-scale, high-resolution annotated RGB images, we developed an automatic Data Generator. This approach eliminates the need for time-consuming manual annotation and ensuring higher ground truth accuracy. The data generator performs random augmentations to particles, like flipping, rotation and colorization, while avoiding scaling to preserve the authenticity of size and shape. The augmented particles are then randomly placed onto a 4096\texttimes 4096 pixels conveyor belt background, simulating various occlusion conditions to enhance both robustness and realism. 
To overcome the storage bottleneck associated with classical binary masks as ground truth, we consolidate all mask annotations in a single portable graymap (PGM) format image. 

\textbf{Multi-stage dataset}: We construct our 4096\text{x}4096 pixels image data in 3 stages. The L1 images consist of instances from a single class, where instances are dropped to cover the entire conveyor belt background as much as possible without any overlapping. 
The L2 images build upon L1 by introducing occlusion, while ensuring that the visibility of each individual object remains within the range of [$60\%$, $100\%$], effectively simulating real industrial scenarios. 
The L3 images further advance to mixed-class, where each image contains instances from different classes. In CDW crushing processes, smaller particles tend to settle to the bottom due to conveyor belt vibrations, while bigger particles remain on the top. To replicate these real-world conditions, we introduce a multi-layer structure to L3 (see Table~\ref{tab:class_layer}), while keeping particles in each layer to maintain a visibility range of [$60\%$, $100\%$]. 
Since PSD (Particle Size Distribution) is a key criterion for analyzing RC materials, we additionally annotate the PSD for all stage images to facilitate more precise analysis. To enhance data diversity and better align with real-world conditions, we incorporate multiple PSD distributions, including uniform distribution, Gaussian distribution, and random distribution. Table \ref{tab:datasets_details} provides an overview of all stages. 

\begin{table}[bt]
\centering
\caption{Categorized Segmented Instances And Layers}
\renewcommand{\arraystretch}{1.2} 
\setlength{\tabcolsep}{6pt} 
\begin{tabular}{ccc|c}
\hline
\textbf{Class} & \textbf{Size Range [mm]} & \textbf{\# Particles} & \textbf{Layer} \\
\hline
Class 1 & 4.0--5.6  & 486 & \multirow{3}{*}{Layer\_0} \\
Class 2 & 5.6--8.0  & 637 &  \\
Class 3 & 8.0--11.2 & 403 &  \\
\cline{4-4}
Class 4 & 11.2--16.0 & 348 & \multirow{2}{*}{Layer\_1} \\
Class 5 & 16.0--22.4 & 333 &  \\
\cline{4-4}
Class 6 & 22.4--35.0 & 185 & Layer\_2 \\
\cline{4-4}
Class 7 & 35.0--45.0 & 106 & Layer\_3 \\
\cline{4-4}
Class 8 & 45.0--63.0 & 8 & Layer\_4 \\
\hline
\end{tabular}
\label{tab:class_layer}
\end{table}

\begin{table}[b]
    \centering
    \caption{Details Of Different Evaluation Datasets}
    \renewcommand{\arraystretch}{1.2}
    \setlength{\tabcolsep}{3pt} 
    \begin{tabular}{c|c c c c c c}
        \hline
        \textbf{\makecell{Data}} & \textbf{\makecell{\# Images}} & \textbf{\makecell{Visibility \\of Particles}} & \textbf{\makecell{\# Particles \\ per Image}} & \textbf{\makecell{Class}} & \textbf{\makecell{Layers}} & \textbf{\makecell{Occlusion}}\\
        \hline
        \text{L}1 & 115$^{\mathrm{a}}$  & 100\%       & [93, 3525]  & single & single & none\\
        \text{L}2\text{-l} & 159$^{\mathrm{a}}$  & 60\%--100\% & [195, 500] & single & single & low\\
        \text{L}2\text{-h} & 121$^{\mathrm{a}}$  & 60\%--100\% & [195, 4831] & single & single & heavy\\
        \text{L}3\text{-0} & 79$^{\mathrm{a}}$  & 0\%--100\%  & [162, 1983] & mixed & single & low\\
        \text{L}3\text{-m} & 239$^{\mathrm{a}}$  & 0\%--100\%  & [698, 3125] & mixed & multi & medium\\
        \text{L}3\text{-h} & 640$^{\mathrm{a}}$  & 0\%--100\%  & [698, 6251] & mixed & multi & heavy\\
        \hline
        \multicolumn{7}{l}{$^{\mathrm{a}}$Resolution of 4096\text{x}4096}
    \end{tabular}
\label{tab:datasets_details}
\end{table}

\subsection{Adapting SAM for Small Particle Segmentation}
\label{sec:adapt}


Despite SAM's strong performance in general instance segmentation, it struggles with complex scenes, especially for small or overlapping objects. Its reliance on prompt inputs limits full automation, posing challenges for our application. By refining its encoder-decoder architecture with multi-stage simulated data, we enhance its ability to segment dense, small particles in high-resolution images.

We adopt SAM’s image encoder and mask decoder to develop a real-time automated segmentation model for aggregates names \textbf{ParticleSAM}. First, unlike the minimum-mask area merging approach in SAM, we introduce a maximum-area filtering to remove the impact from the conveyor belt background and improve inference speed, particularly for segmenting fine-crushed particles. Next, we focus on two aspects of adaptation:

\textbf{Feature Extraction Enhancement}: In SAM’s everything demo, features are extracted using a 32×32 grid of sampling points. To enhance feature representation of our 4096\text{x}4096 pixel images, we increase this to a 64×64 grid. Additionally, we use the image splitting functionality to split input image into four crops, allowing the model to extract local features before final feature fusion. While this approach enriches features encoding, it also increases inference time by over 5 times.  To achieve real-time monitoring in the application, we choose the smallest backbone of SAM (\text{ViT-B}) and no sub-crops to form our ParticleSAM.

\textbf{Postprocessing Optimization}: We tune the post-processing parameters, including Non-Maximum Suppression (NMS), Intersection over Union (IoU) thresholds, and stability constraints, to efficiently remove padding and filter duplicate masks, thereby enhancing the decoder's efficiency. This optimization improves segmentation precision by reducing false positives and ensuring greater consistency, making the results better aligned with our industrial application.

\begin{figure}[ht]
    \centering
    \includegraphics[width=0.45\textwidth]{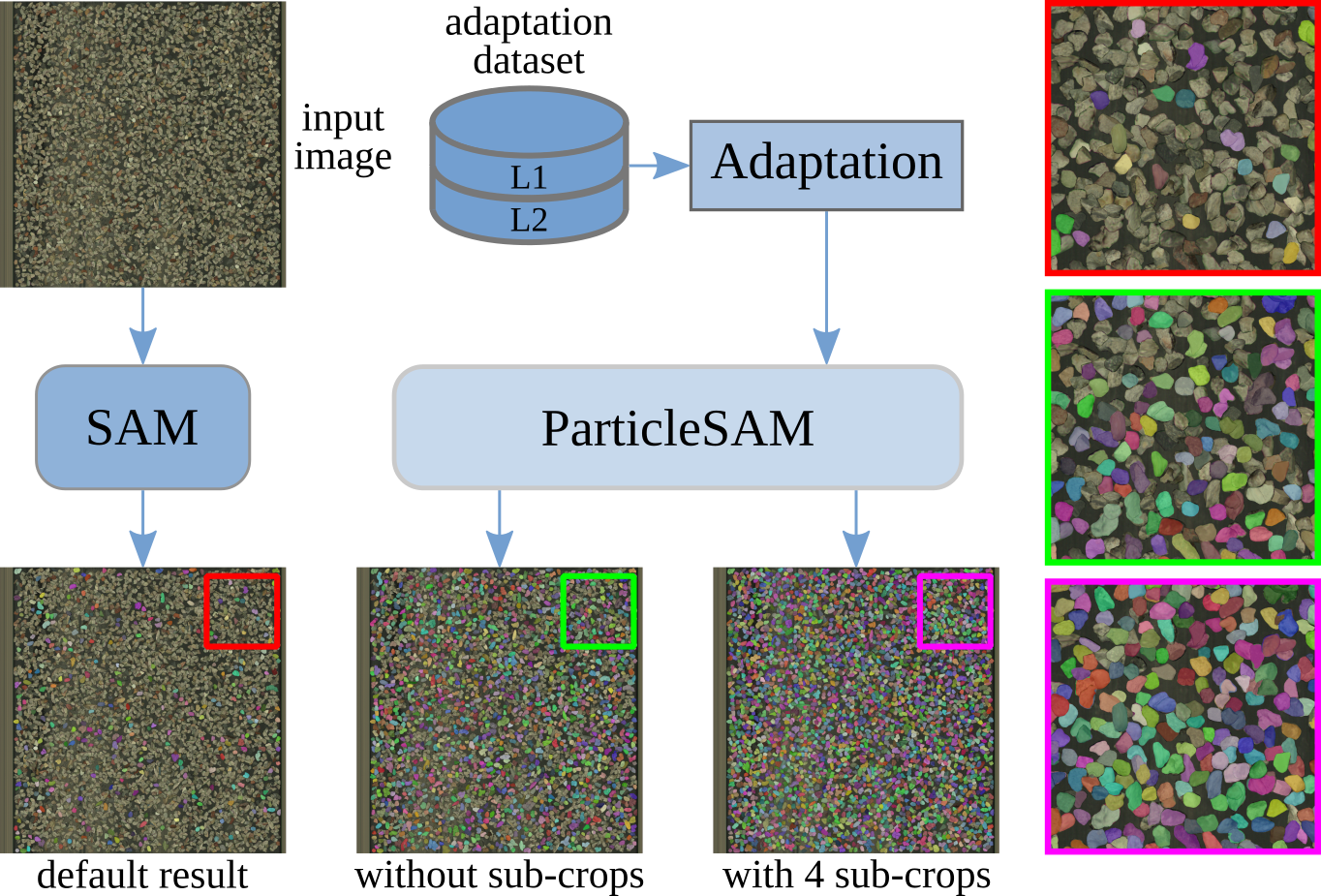}  
    \caption{Overlay of input images with colored segmentation masks shwoing the improvement of particle segmentation of ParticleSAM compared to SAM with 4096x4096 image data. Images on the right are zoomed-in crops.
    \textit{The source conveyor belt background provided by the department of Anthropogenic Material Cycles (ANTS), RWTH Aachen University.}}
    \label{Figure 2}
\end{figure}

Last, to systematically optimize the model's performance, we use a progressive hyperparameter tuning strategy on our simulated data. 
An adaptation subset of images is randomly selected from both the L1 and L2 data. The tuning process begins with the subset of no-occlusion data (L1), and parameters are optimized in descending order of particle size. Similarly to the L1 phase, adaptation starts with the largest class in L2 and gradually moves to smaller classes. Fig.~\ref{Figure 2} depicts the different adapted results.

\section{Experiments}

\textbf{Metrics}: We report the mean intersection-over-union (mIoU) metric to quantify the accuracy of particle segmentation. We also use the mean average precision (mAP) metric with thresholds to evaluate the model's ability to recognize particles, e.g. mAP\textsubscript{50} considers a segmentation is correct when the IoU between the predicted and the groundtruth masks is larger than 50\%. 

\textbf{Baseline}: Since we are not aware of other particles segmentation research for crushed waste, we leverage SAM’s Automatic Mask Generator (AMG) as a baseline to compare our ParticleSAM's performance. Our ParticleSAM is adapted from the lightweight backbone of SAM (\text{ViT-B}), and the AMG retains the same \text{ViT-B} backbone.

\textbf{Evaluation Data}: Excluding the subset of L1 and L2 images used for adaptation, the remaining images are utilized to evaluate the model’s performance. Table~\ref{tab:datasets_details} provides an overview of all the test datasets. Additionally, to assess the model’s improvement in segmenting extremely small objects, we generate extra L3-0 data, which contains only layer\textsubscript{0} and layer\textsubscript{1} particles with size ranging from 4 to 11.2 \text{mm}. Moreover, to compare performance under occlusion, we partition L2 and L3 data into less occluded and heavily occluded. Specifically, due to the annotations of PSD per image, the less occluded image keeps the same PSD as the heavily occluded images, but the number of stones per class is reduced by 50\%. 

\textbf{Discussion}: Table~\ref{tab:model_results} shows segmentation performance of SAM and our ParticleSAM with various metrics from different evaluation data. ParticleSAM consistently outperforms SAM across all datasets. Notably, on the occluded mixed-class L3 data, SAM fails to deliver meaningful results, whereas ParticleSAM demonstrates a remarkable improvement of 26.09\% and 50.30\% on mIoU.
Moreover, on the L3-0 dataset, which consists exclusively of particles smaller than 22 mm, by comparing mAP under different thresholds, we can observe that our model not only improves accuracy but also exhibits more stable segmentation performance with less drop from mAP\textsubscript{50} to mAP\textsubscript{90}.

\begin{figure}[ht]
    \centering
    \includegraphics[width=0.4\textwidth]{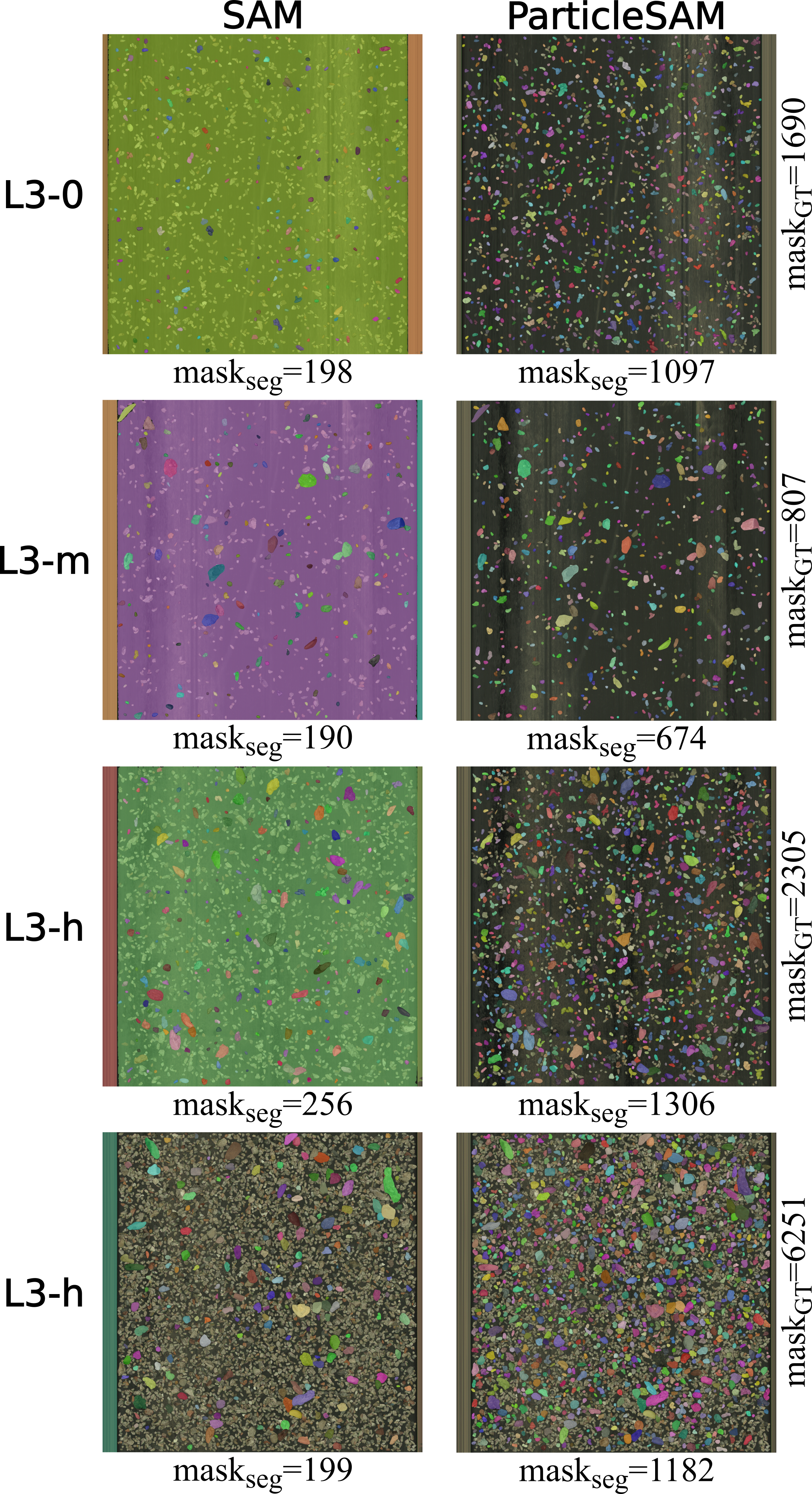}  
    \caption{Comparison of SAM and ParticleSAM segmentation across L3 dataset occlusion levels, with the number of segmented vs. total GT masks shown.
     \textit{The source conveyor belt background provided by the department of Anthropogenic Material Cycles (ANTS), RWTH Aachen University.}}
    \label{Figure 3}
\end{figure}

Although our model achieves a significant improvement in particle segmentation, its performance declines under occlusion. Comparison in L2-l and L2-h, L3-m and L3-h, heavily occlusion leads to a noticeable performance drop, from 85.53\% to 45.15\% and 69.25\% to 34.79\%, respectively. In the most heavily occluded dataset L3-h, ParticleSAM's detection capability is limited as the number of particles increases, failing to recognize more particles (See Fig.~\ref{Figure 3}). As an alternative, sub-cropping, discussed in Section \ref{sec:adapt}, can mitigate this limitation and detect more particles under heavy occlusion; however, this approach increases inference time more than 5 times, making real-time monitoring impractical.

\begin{table}[t]
    \centering
    \caption{Particle Segmentation Comparisons Between SAM And Our ParticleSAM On Different Datasets}
    \label{tab:model_results}
    \renewcommand{\arraystretch}{1.2}
    \setlength{\tabcolsep}{4pt} 
    \begin{tabular}{cccccccc}
        \hline
        \multirow{2}{*}{\textbf{Data}} & \multirow{2}{*}{\textbf{Model}} & \multirow{2}{*}{\textbf{mIoU}} & \textbf{mAP\textsubscript{50}}$^{\mathrm{a}}$ & \textbf{mAP\textsubscript{60}} & \textbf{mAP\textsubscript{70}} & \textbf{mAP\textsubscript{80}} & \textbf{mAP\textsubscript{90}}\\
        \cline{4-8}
        & & & (\%) & (\%) & (\%) & (\%) & (\%) \\
        \hline
        \multirow{2}{*}{L1} & SAM & 55.80 & 53.60 & 46.53 & 43.27 & 42.18  & 41.23\\
        & \textbf{Ours} & \textbf{68.17} & \textbf{61.12} & \textbf{54.26} & \textbf{50.83} & \textbf{49.51} & \textbf{48.51} \\
        \hline
        \multirow{2}{*}\text{L}2\text{-l} & SAM & 61.09 & 73.16 & 72.50 & 71.87 & 70.62 & 65.53 \\
        & \textbf{Ours} & \textbf{85.53} & \textbf{89.64} & \textbf{89.32} & \textbf{88.86} & \textbf{87.98} & \textbf{85.72} \\
        \hline
        \multirow{2}{*}\text{L}2\text{-h} & SAM & 27.35 & 38.82 & 33.39 & 29.59 & 25.70  & 21.22\\
        & \textbf{Ours} & \textbf{45.15} & \textbf{48.10} & \textbf{42.65} & \textbf{38.33} & \textbf{33.97} & \textbf{29.11} \\
        \hline
        \multirow{2}{*}\text{L}3\text{-0} & SAM & 18.26 & 34.23 & 32.48 & 31.30 & 30.33 & 27.65 \\
        & \textbf{Ours} & \textbf{44.13} & \textbf{47.65} & \textbf{45.90} & \textbf{44.70} & \textbf{43.69} & \textbf{41.98} \\
        \hline
        \multirow{2}{*}\text{L}3\text{-m} & SAM & 18.95 & 55.23 & 53.95 & 52.77 & 51.31  & 44.67\\
        & \textbf{Ours} & \textbf{69.25} & \textbf{82.78} & \textbf{82.03} & \textbf{81.26} & \textbf{80.27}  & \textbf{76.88}\\
        \hline
        \multirow{2}{*}\text{L}3\text{-h} & SAM & 8.70 & 30.06 & 28.25 & 26.80 & 25.32 & 21.87 \\
        & \textbf{Ours} & \textbf{34.79} & \textbf{43.24} & \textbf{41.19} & \textbf{39.46} & \textbf{37.79} & \textbf{34.98} \\
        \hline
        \multicolumn{8}{l}{$^{\mathrm{a}}$mAP\textsubscript{50} represent IoU threshold at 50\%}
    \end{tabular}
\end{table}

\section{Conclusion}
We propose ParticleSAM, an adaptation of SAM for particle segmentation in high-resolution images with numerous small and dense objects.
The challenge is addressed by parameter adaptation, the encoder of SAM to handle larger images and extract features on a smaller scale and the decoder to optimize the segmentation precision. 
%
Moreover, to the best of our knowledge, we are the first to introduce a large-scale simulated dataset in recycled CDW, specifically designed for crushed aggregates. This dataset provides a novel benchmark for fine-tuning and evaluating segmentation models in challenging scenarios as high-density occlusions or small particle material analysis, bridging a critical gap in existing datasets. 
The results on our dataset demonstrate a significant enhancement in particle detection and segmentation performance compared to SAM, especially on images with overlapping particles. 
%
Although our work was intended for the visual quality inspection of CDW, our proposed method ParticleSAM can be applied to other domains, where visual monitoring requires the segmentation of multiple, small or overlapping objects, in which case our work can serve as a guideline. 
Similarly, our data generation pipeline for multi-layer, overlapping particle images can be easily adapted to any type of object, e.g. in medical and agriculture applications. 

\section*{Acknowledgment}
We would like to thank the Department of Anthropogenic Material Cycles, RWTH Aachen University, for providing the raw images and backgrounds for the synthetic images, recorded at their facility.

\balance
\bibliography{refs}

\begin{thebibliography}{10}
\providecommand{\url}[1]{#1}
\csname url@samestyle\endcsname
\providecommand{\newblock}{\relax}
\providecommand{\bibinfo}[2]{#2}
\providecommand{\BIBentrySTDinterwordspacing}{\spaceskip=0pt\relax}
\providecommand{\BIBentryALTinterwordstretchfactor}{4}
\providecommand{\BIBentryALTinterwordspacing}{\spaceskip=\fontdimen2\font plus
\BIBentryALTinterwordstretchfactor\fontdimen3\font minus \fontdimen4\font\relax}
\providecommand{\BIBforeignlanguage}[2]{{%
\expandafter\ifx\csname l@#1\endcsname\relax
\typeout{** WARNING: IEEEtran.bst: No hyphenation pattern has been}%
\typeout{** loaded for the language `#1'. Using the pattern for}%
\typeout{** the default language instead.}%
\else
\language=\csname l@#1\endcsname
\fi
#2}}
\providecommand{\BIBdecl}{\relax}
\BIBdecl

\bibitem{aggregatesEurope}
\BIBentryALTinterwordspacing
{Aggregates Europe}, ``Facts and figures,'' 2025, accessed: 2025-03-13. [Online]. Available: \url{https://www.aggregates-europe.eu/facts-figures/facts/}
\BIBentrySTDinterwordspacing

\bibitem{kirillov2023segment}
A.~Kirillov, E.~Mintun, N.~Ravi, H.~Mao, C.~Rolland, L.~Gustafson, T.~Xiao, S.~Whitehead, A.~C. Berg, W.-Y. Lo \emph{et~al.}, ``Segment anything,'' in \emph{Proceedings of the IEEE/CVF international conference on computer vision}, 2023, pp. 4015--4026.

\bibitem{hernandez2021near}
J.~Hern{\'a}ndez~Parrodi, N.~Kroell, X.~Chen, T.~Dietl, E.~Pfund, B.~K{\"u}ppers, and C.~Nordmann, ``Near-infrared-based material flow monitoring of construction and demolition waste nahinfrarot-basierte strom{\"u}berwachung von bau-und abbruchabf{\"a}llen,'' \emph{Mineralische Nebenprodukte Und Abf{\"a}lle. Berlin: Thom{\'e}-Kozmiensky Verlag GmbH}, pp. 92--111, 2021.

\bibitem{dlPSD}
N.~Kroell, E.~Thor, L.~G{\"o}bbels, P.~Sch{\"o}nfelder, and X.~Chen, ``Deep learning-based prediction of particle size distributions in construction and demolition waste recycling using convolutional neural networks on 3d laser triangulation data,'' \emph{Construction and Building Materials}, vol. 466, p. 140214, 2025.

\bibitem{rgbd}
J.~Li, H.~Fang, L.~Fan, J.~Yang, T.~Ji, and Q.~Chen, ``Rgb-d fusion models for construction and demolition waste detection,'' \emph{Waste Management}, vol. 139, pp. 96--104, 2022.

\bibitem{CDWdatasets}
T.~Ji, J.~Li, H.~Fang, R.~Zhang, J.~Yang, and L.~Fan, ``Rapid dataset generation methods for stacked construction solid waste based on machine vision and deep learning,'' \emph{Plos one}, vol.~19, no.~1, p. e0296666, 2024.

\bibitem{doadaptSeg}
S.~K. Dodampegama, L.~Hou, E.~Asadi, K.~Zhang, and S.~Setunge, ``Adversarial domain adaptation with a modified yolov8x-seg model for construction and demolition waste segmentation,'' 2024, preprint available at SSRN 5106445.

\bibitem{wastesam}
S.~Heo and S.~Na, ``Developing wastesam: A novel approach for accurate construction waste image segmentation to facilitate efficient recycling,'' \emph{Waste Management \& Research}, p. 0734242X241290743, 2024.

\bibitem{kronenwett2024sensor}
F.~Kronenwett, G.~Maier, N.~Leiss, R.~Gruna, V.~Thome, and T.~L{\"a}ngle, ``Sensor-based characterization of construction and demolition waste at high occupancy densities using synthetic training data and deep learning,'' \emph{Waste Management \& Research}, vol.~42, no.~9, pp. 788--796, 2024.

\bibitem{mask2former}
B.~Cheng, I.~Misra, A.~G. Schwing, A.~Kirillov, and R.~Girdhar, ``Masked-attention mask transformer for universal image segmentation,'' in \emph{Proceedings of the IEEE/CVF conference on computer vision and pattern recognition}, 2022, pp. 1290--1299.

\bibitem{das2024prompting}
A.~M. Das, R.~Chaudhry, K.~Kundu, and D.~Modolo, ``Prompting foundational models for omni-supervised instance segmentation,'' in \emph{Proceedings of the IEEE/CVF Conference on Computer Vision and Pattern Recognition}, 2024, pp. 1583--1592.

\bibitem{wu2023medical}
J.~Wu, W.~Ji, Y.~Liu, H.~Fu, M.~Xu, Y.~Xu, and Y.~Jin, ``Medical sam adapter: Adapting segment anything model for medical image segmentation,'' \emph{arXiv preprint arXiv:2304.12620}, 2023.

\bibitem{CDWreview}
A.~Langley, M.~Lonergan, T.~Huang, and M.~R. Azghadi, ``Analyzing mixed construction and demolition waste in material recovery facilities: Evolution, challenges, and applications of computer vision and deep learning,'' \emph{Resources, Conservation and Recycling}, vol. 217, p. 108218, 2025.

\bibitem{DIN}
X.~Wu, N.~Kroell, and K.~Greiff, ``Deep learning-based instance segmentation on 3d laser triangulation data for inline monitoring of particle size distributions in construction and demolition waste recycling,'' \emph{Resources, Conservation and Recycling}, vol. 205, p. 107541, 2024.

\end{thebibliography}
\bibliographystyle{IEEEtran}

\end{document}